\documentclass[sigconf,natbib=true,anonymous=false,pbalance,screen]{acmart}

\copyrightyear{2026}
\acmYear{2026}
\setcopyright{cc}
\setcctype{by}
\acmConference[CIKM '26]{Proceedings of the 35th ACM International Conference on Information and Knowledge Management}{November 07--11, 2026}{Rome, Italy}
\acmBooktitle{Proceedings of the 35th ACM International Conference on Information and Knowledge Management (CIKM '26), November 07--11, 2026, Rome, Italy}
\acmDOI{10.1145/3799682.3839850}
\acmISBN{979-8-4007-2539-5/2026/11}

\usepackage{booktabs}
\usepackage{bbm}
\usepackage{amsthm}
\usepackage{graphicx}
\usepackage{algorithm}
\usepackage{algpseudocode}
\usepackage{enumerate}

\newtheorem{definition}{Definition}
\newtheorem{proposition}{Proposition}

\floatname{algorithm}{Procedure}

\begin{document}

\title{BIRDNet: Mining and Encoding Boolean Implication Knowledge Graphs as Interpretable Deep Neural Networks}

\author{Tirtharaj Dash}
\orcid{0000-0001-5965-8286}
\affiliation{%
  \institution{BITS Pilani, K K Birla Goa Campus}
  \city{Zuarinagar}
  \state{Goa 403726}
  \country{India}
}
\email{tirtharaj@goa.bits-pilani.ac.in}

\renewcommand{\shortauthors}{T. Dash}

\begin{abstract}
Tabular data in knowledge-rich domains often carries a latent prior in the form of Boolean implication relationships (BIRs) between pairs of features. 
We mine such relationships with a sparse-exception binomial test. 
We encode the resulting typed graph as the connectivity of a layered neural network, called BIRDNet, in which 
each hidden unit corresponds to one mined rule and 
binds only to its two features. 
We show two consequences of this design:
First, the architecture is sparse: at most $2/d$ of the weights in each BIR layer are active, where $d$ is the input dimension. Second, the model is intrinsically interpretable: every trained unit keeps a stable symbolic identity, so rules can be read off the network without surrogate models. 
Unlike most neurosymbolic models, BIRDNet does not consume an external rule base; its structural prior is mined from the data. 
We evaluate BIRDNet on six transcriptomic and proteomic benchmarks. 
Our results show that BIRDNet stays within $0.02$ AUROC of the strongest dense baseline, while using up to $95\times$ fewer active parameters than an architecture-matched dense MLP. 
First-layer rules align with known biological signatures across multiple
cancer subtypes and tissue types. 
Matched-topology controls show that the mined graph contributes symbolic
meaning rather than predictive advantage: shuffled or random pairings
match or improve AUROC but no longer correspond to mined implications.
Data and code are available at: \url{https://github.com/MAHI-Group/BIRDNet}.
\end{abstract}



\maketitle

\section{Introduction}

Tabular data in knowledge-rich scientific domains often carries latent
symbolic structure that a black-box predictor cannot fully exploit. 
In gene
expression and proteomics, and more broadly across binarisable
measurement modalities, pairs of features routinely satisfy strong
logical relationships of the form 
``high $a$ implies high $b$,''
``low $a$ implies low $b$,'' and their negations~\cite{sahoo2008}.
The complete set of such relationships across all feature pairs
constitutes a typed directed graph that can be mined directly from data
without supervision.
Its strongest implications often
align with known relational structure in the domain of interest
such as
gene-regulatory interactions in transcriptomics or protein interactions in proteomics datasets~\cite{sahoo2012power}.

In this paper, we investigate
whether this mined graph can serve as the \emph{architecture}
of a deep neural network. 
To this end, we propose BIRDNet, a network whose connectivity
in each hidden layer is determined entirely by an implication graph
mined from available training data. 
Each hidden unit in a BIRDNet corresponds to one mined implication. It connects only to the two features participating in that implication, and a binary mask fixed at construction time enforces this structure throughout training.
While the first hidden layer of BIRDNet corresponds directly to
implications mined among input features, defining the connectivity
of subsequent layers is less direct: there are no raw features to
mine over. We address this by constructing subsequent layers
greedily: layer $\ell$
mines a fresh implication graph on the
post-activation outputs of layer $\ell{-}1$, and contributes one
hidden unit per mined implication. 
Sparsity follows from the same
topological constraint at every layer, since each hidden unit has
exactly two active incoming weights.
The same constraint gives each hidden unit a stable symbolic identity.
First-layer units correspond directly to implications over named input
features, while deeper units relate earlier rules. This hierarchy can be
traced back to the inputs, with LRP used to identify units relevant to a
prediction.

\subsubsection*{Related work.}
Building neurosymbolic AI models typically relies on an external, hand-curated rule base or ontology, with a neural model constrained to respect it. For instance, in biomedicine, DCell~\cite{ma2018dcell} encodes the Gene Ontology and P-NET~\cite{elmarakeby2021pnet} encodes Reactome pathways, in both cases routing connectivity through a curated hierarchy of biological concepts. Outside biology, Compositional Relational Machines~\cite{srinivasan2024crm} populate vertices with first-order features composed via an expert-specified mode language, and a broader family of approaches injects logical rules or knowledge graphs into neural training as soft constraints or differentiable layers~\cite{garcez2023neurosymbolic,wang2024towards,dash2022review}. In all of these, the symbolic prior is external to the data, and the role of the network is to obey it. BIRDNet takes a different route. Its structural prior is not supplied; it is mined from the training data by a statistical test with a precise propositional reading. The symbolic content the network internalises is therefore already present in the data it is asked to model, rather than imposed on top of it. Of course, this data-driven prior has its own limitations, which we discuss in Section~\ref{sec:concl}.

Our formulation of the structural prior follows the Boolean implication framework of Sahoo et al.~\cite{sahoo2008,sahoo2012power}, which introduced the StepMiner binarisation and the sparse-exception binomial test we use here.
In that line of work, the implication network is treated as an analytical object, mined and used to select gene sets that are then passed to a separate downstream model.
Here we keep the mining machinery and change what the graph is used for: the mined graph becomes the structure of a deep network, with implication edges as its hidden-layer connectivity.

\subsubsection*{Contributions.}
Our contributions are the following:
\begin{enumerate}[(1)]
\item we formalise Boolean implication knowledge graphs as a typed, data-mineable representation suitable for use as a structural prior of a deep neural network;
\item we introduce BIRDNet, a layer-wise sparse architecture that encodes the mined graph as connectivity, prove a $2/d$ upper bound on the active-weight fraction per BIR layer (so the compression ratio over an architecture-matched dense layer grows linearly with input dimension $d$), 
and give a rule-extraction procedure that maps trained units back to weighted symbolic rules; and
\item we evaluate BIRDNet on six biomedical benchmarks spanning
transcriptomics and proteomics.
Beyond the main results, implementation details, per-class rule tables for all six datasets, and per-instance rule traces are made available in a supplementary file in the code repository.
\end{enumerate}

\section{Method}
\label{sec:method}

\subsection{Mining the implication knowledge graph}
\label{sec:method-mining}

Let $X \in \mathbb{R}^{n \times d}$ be a feature matrix over $n$ samples and $d$ features. Each feature is binarised independently using the StepMiner threshold $\tau_j$, 
obtained by sorting the values of $X_{\cdot j}$ and fitting a single-step function that minimises the within-segment sum of squared residuals~\cite{sahoo2008}. 
The step location $\tau_j$ separates a low-valued segment from a high-valued segment, 
which suits features with approximately bimodal distributions. 
The binarised matrix $B \in \{0,1\}^{n \times d}$ has $B_{ij} = \mathbf{1}[X_{ij} > \tau_j]$.

A Boolean implication relationship (BIR) is defined over an ordered pair of features $(a, b)$. 
For each such pair we test four primary implication types $T_0\!:a^H{\to}b^H$, $T_1\!:a^L{\to}b^L$, $T_2\!:a^H{\to}b^L$, $T_3\!:a^L{\to}b^H$, where $a^H, a^L$ denote $B_{\cdot a}=1$ and $B_{\cdot a}=0$ respectively. 
For each candidate we count exception samples (those violating the
implication) and test the null that this count is consistent with a
binomial whose success probability is the marginal product under
independence. 
An implication is asserted whenever the left-tail
$p$-value falls below $p^* = 10^{-6}$, that is, when the observed
number of exceptions is significantly \emph{smaller} than independence
predicts, and the exception fraction does not exceed
$\pi = 0.05$~\cite{sahoo2008}. We restrict candidates to features whose
binarised marginal lies in $[0.05, 0.95]$, so that neither literal is
near-constant. 
Pairs satisfying both
$T_0$ and $T_1$ are labelled \emph{equivalent} ($T_4$, $a \equiv b$);
pairs satisfying both $T_2$ and $T_3$ are labelled \emph{opposite}
($T_5$, $a \equiv \neg b$). The output is a typed directed graph
$\mathcal{G}=(V,E,\mathit{type})$ with $\mathit{type}:E\to\{T_0,\dots,T_5\}$.
We call $\mathcal{G}$ the \emph{Boolean implication knowledge graph} of $X$.

\subsubsection*{Propositional semantics.}
For each feature $a$, define the Boolean atom $A := (X_a > \tau_a)$, 
and similarly $B$ for feature $b$.
The four primary BIR types correspond to the two-literal clauses
$A{\to}B$, $\lnot A{\to}\lnot B$, $A{\to}\lnot B$, and $\lnot A{\to}B$,
each holding in $\ge\!1{-}\pi$ of the data and associated with a $p$-value. 
Composite types $T_4$ and $T_5$ are
conjunctions of two such clauses, giving $A{\leftrightarrow}B$ and
$A{\leftrightarrow}\lnot B$; only the primary types are instantiated as
units, so the rule base BIRDNet encodes as its structure consists of
two-literal clauses throughout.

\subsubsection*{Relation to frequent-itemset mining.}
Our mining procedure differs from Apriori and FP-Growth~\cite{agrawal1994fast,han2000mining}, which first identify
co-occurring itemsets above a minimum-support threshold and then derive
association rules from them. In contrast, BIRs are typed by implication
direction and polarity and are selected using a binomial test on the number
of \emph{exceptions}, together with an exception-rate constraint. Thus, a
low-support pair may still yield a BIR when its number of exceptions is
significantly smaller than expected under independence. We report lift
alongside precision in Section~\ref{sec:rules} as a common descriptive
measure; a direct comparison with support-based rule mining is left to
future work.

\subsection{Encoding \texorpdfstring{$\mathcal{G}$}{G} as a deep neural network}
\label{sec:method-arch}

\begin{figure}[!htb]
\includegraphics[width=\linewidth]{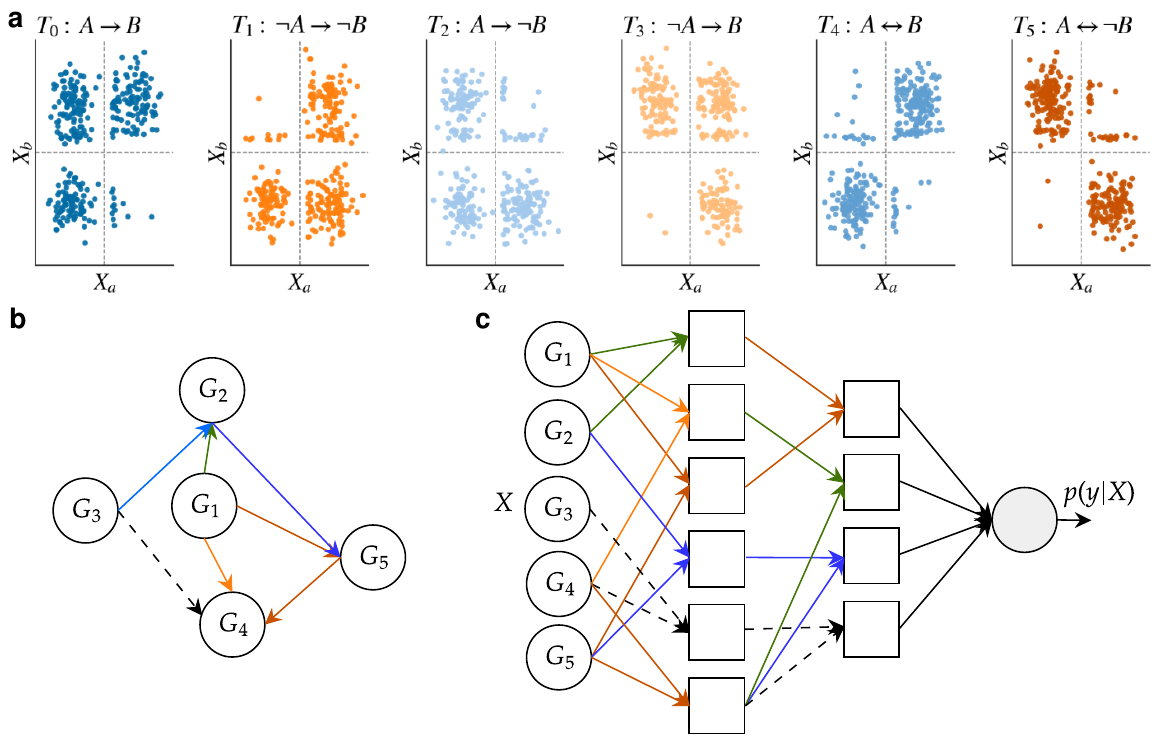}
\caption{BIRDNet construction. 
(a) The six BIR types as binarised-quadrant patterns; 
(b) a fragment of the mined Boolean implication knowledge graph; 
(c) the graph encoded as network connectivity, with each hidden unit binding exactly two predecessors. 
A small dense classifier head (omitted) projects the final BIR-layer activations onto class logits.}
\label{fig:overview}
\Description{Overview of the BIR types, knowledge graph, and network connectivity.}
\end{figure}

In Figure~\ref{fig:overview} we show the six types of BIRs, a knowledge graph fragment, and its encoding as a BIR layer for a deep network.

\begin{definition}[BIR layer]
\label{def:birlayer}
Given a feature space of dimension $d$ and a set of mined implications
$\mathcal{B} = \{(i_k, j_k, t_k)\}_{k=1}^{h}$, the \emph{BIR layer}
$f_{\mathcal{B}}: \mathbb{R}^d \to \mathbb{R}^{h}$ has weight matrix
$W \in \mathbb{R}^{h \times d}$ and a fixed binary mask
$M \in \{0,1\}^{h \times d}$ with $M_{k,m} = 1$ iff
$m \in \{i_k, j_k\}$. A forward pass computation is then
$f_{\mathcal{B}}(x) = (W \odot M)x + b$ where $\odot$ is
elementwise multiplication.
\end{definition}

The mask is applied at every forward pass, so the gradient with
respect to any $W_{k,m}$ with $M_{k,m}{=}0$ is exactly zero. 
The sparsity pattern therefore persists throughout training as a hard structural constraint.
In many scientific domains that we are interested in, usually $d \sim 10^3$ to $10^4$. 
Proposition~\ref{prop:sparsity} below provides a bound that predicts $>\!99.8\%$ of BIR-layer weights are zeroed by mask alone, before any learned sparsity such as regularisation or pruning~\cite{frankle2019lottery,xia2022structured}.
Each mined implication is represented by a single hidden unit, whose two active weights are initialised according to the implication type. For ($T_0$), both weights are initialised positively; for ($T_1$), both are initialised negatively. For ($T_2$), the source weight is positive and the target weight negative, while ($T_3$) uses the opposite configuration. Composite types ($T_4$) and ($T_5$) are retained during the mining stage but are not instantiated as separate hidden units. Instead, a composite type is considered present only when both of its constituent primary types are present. To avoid redundant representations, at most one hidden unit is instantiated for each feature pair.

\begin{proposition}[BIR-Layer Sparsity]
\label{prop:sparsity}
Consider a BIR layer encoding $h$ implications over $d$ input features.
By Definition~\ref{def:birlayer}, each row of $W \odot M$ contains at
most two non-zero entries. Hence, $\|W \odot M\|_0 \leq 2h$,
compared with $hd$ weights in the corresponding dense layer. Therefore,
the fraction of active weights is at most $2/d$, independent of the
number of implications $h$. This bound applies specifically to the BIR
layers; the overall compression relative to a dense baseline is limited
by the shared classifier head, as discussed in
Section~\ref{sec:experiments}.
\end{proposition}

\subsubsection*{Bounded in-degree as the structural prior.}
BIRDNet shares a bounded-in-degree prior with Compositional Relational
Machines~\cite{srinivasan2024crm}, where each non-input vertex composes at
most two predecessors. In CRM, admissible first-order relational features
are specified through a mode language; in BIRDNet, pairwise Boolean
implications are discovered from data by a statistical test. Thus, the
architectural constraint is similar, but the symbolic structure is
data-mined rather than supplied as background knowledge.

\subsubsection*{Greedy layer-wise construction.}
The depth of BIRDNet is capped at a maximum value $L$ but not fixed in advance: construction stops early as soon as a fresh layer produces fewer than $\mu$ valid implications. 
Layer $0$ mines $\mathcal{G}_0$ on the input and contributes one unit per mined implication. We pass its outputs through BatchNorm, ReLU, and dropout, and layer $\ell{+}1$ then mines a fresh graph $\mathcal{G}_{\ell+1}$ on these post-activation outputs. We append a dense classifier head that maps the final BIR-layer outputs to class logits. 
We illustrate this procedure in Procedure~\ref{alg:greedy}.
A vectorised parallel
implementation of the mining step is included in the released code.

\begin{algorithm}[!htb]
\caption{Greedy layer-wise construction of BIRDNet.}
\label{alg:greedy}
\begin{algorithmic}[1]
\State \textbf{Input:} training data $X$; thresholds $p^*, \pi$; cap $h_{\max}$; floor $\mu$; max depth $L$.
\State $H \gets X$
\For{$\ell = 0, \dots, L-1$}
  \State $\tilde{H} \gets \textsc{StepMinerBinarise}(H)$
  \State $\mathcal{G}_\ell \gets \textsc{MineBIRs}(\tilde{H}, p^*, \pi)$
  \State $\mathcal{B}_\ell \gets \textsc{Deduplicate}(\mathcal{G}_\ell)$; keep top-$h_{\max}$ by $p$-value
  \If{$|\mathcal{B}_\ell| < \mu$} \State \textbf{break} \EndIf
  \State Append BIR layer $f_{\mathcal{B}_\ell}$ + BN + ReLU + dropout to $f$
  \State $H \gets$ post-activation outputs of $f$ on $X$
\EndFor
\State Append a dense classifier head to $f$
\State \textbf{Return} $f$
\end{algorithmic}
\end{algorithm}

\subsection{Reading rules from a trained BIRDNet}
\label{sec:method-rules}
Each unit $k$ in the first hidden-layer corresponds to one mined implication
$(i,j,t)$. 
For each class $c$ we estimate $\Pr(c \mid k \text{ active})$
on a held-out set (a unit is active when its post-ReLU output is
positive) and report precision, recall, lift (the ratio
$\Pr(c \mid k \text{ active}) / \Pr(c)$, where $1$ denotes
independence and values $> 1$ indicate class enrichment), and
support. 
We rank each rule by its associated precision as the primary metric and report
lift alongside, since precision alone can be inflated by class
prevalence while remaining an intuitive metric for domain-experts. 
Per-instance explanations are obtained by Layer-wise
Relevance Propagation~\cite{bach2015lrp}: relevance flows back to
individual BIR units, each of which encodes a propositional rule from mined implications.
This allows a given prediction to be explained directly in a hierarchical manner (as in~\cite{srinivasan2024crm,srinivasan2019logical}), without requiring a surrogate model.

\section{Empirical Evaluation}
\label{sec:experiments}

\subsection{Datasets and setup}

We evaluate BIRDNet on six benchmarks from two domains: transcriptomics and proteomics (Table~\ref{tab:datasets}). 
For the high-$d$ transcriptomic datasets we apply ANOVA F-test preselection to retain the top $2{,}000$ features. The selection is fitted on each training fold separately, so no test label influences the feature set. For TCGA RNA-seq, the dataset is pre-reduced offline to the $5{,}000$ most-variable genes, an unsupervised step, before per-fold F-test preselection to $2{,}000$ features.
TCGA RPPA and UCI mice protein are kept at full feature width, and so involve no preselection at all.
We report a label-free selection control in our ablation studies.
We use 5-fold stratified cross-validation, with a further $15\%$ of each training fold held out for early stopping.
Features are standardised on the training fold,
and BIRs are mined on the training fold only.

\begin{table}[!htb]
\caption{Datasets used in our experiments. $n$ = number of samples;
  $d_{\text{raw}}$ = number of features in the original dataset;
  $d$ = number of features used after preselection;
  $k$ = number of classes. T = transcriptomics, P = proteomics.}
\label{tab:datasets}
\centering
\small
\begin{tabular}{lcrrrr}
\toprule
Dataset           & Mod. & $n$       & $d_{\text{raw}}$ & $d$    & $k$ \\
\midrule
UCI mice protein~\cite{higuera2015self}     & P & 1{,}080 &   77    &   77    &  8 \\
TCGA RPPA~\cite{akbani2014pan}         & P & 7{,}500 &  258    &  258    & 27 \\
GSE39582~\cite{marisa2013gene}          & T &    566  & 54{,}675 & 2{,}000 &  6 \\
UCI gene expr.~\cite{fiorini2016uci}        & T &    801  & 20{,}531 & 2{,}000 &  5 \\
METABRIC~\cite{curtis2012genomic}          & T & 1{,}974 & 20{,}384 & 2{,}000 &  6 \\
TCGA RNA-seq~\cite{weinstein2013cancer}       & T & 10{,}051 & 5{,}000 & 2{,}000 & 26 \\
\bottomrule
\end{tabular}
\end{table}

We compare BIRDNet against three baselines:
(1) \emph{MatchedMLP},  a fully-dense MLP counterpart of BIRDNet with identical depth, 
layer widths,
batch norm, activation, dropout, classifier head, and training configuration
(AdamW, cosine annealing, gradient clipping, early stopping); that is, only
the BIR mask is removed;
(2) \emph{Logistic Regression with $L_1$
penalty}, a sparse linear baseline; and
(3) \emph{Random Forest}, a
non-linear baseline insensitive to feature scale. 
For BIR mining, we use the following hyperparameters: 
$p^* = 10^{-6}$, $\pi = 0.05$, per-layer cap $h_{\max} = 5{,}000$, 
BIR-count floor $\mu = 10$, maximum depth $L = 2$.
All runs use fixed seed 42 with fully deterministic setup in PyTorch.

\subsection{Results}
\label{sec:results}

\subsubsection{Predictive evaluation}

Table~\ref{tab:results} reports 5-fold AUROC and accuracy.
We treat AUROC as our primary metric because it measures class ranking
independently of the choice of decision threshold, making it particularly
appropriate for comparing a capacity-constrained model across balanced
(UCI gene expr.) and highly imbalanced (TCGA pan-cancer) settings.
BIRDNet's AUROC differs by at most $0.02$ from the strongest dense baseline
on all six datasets, and by at most $0.005$ on TCGA RPPA, UCI mice protein,
UCI gene expr., and TCGA RNA-seq. On TCGA RPPA, BIRDNet attains the highest
AUROC of any method. 
The accuracy gap is larger, ranging from less than one percentage point on
TCGA RPPA and UCI gene expr. to $4.6$ points on METABRIC, $5.5$ points on
TCGA RNA-seq, and $6.5$ points on GSE39582.
The larger discrepancy in accuracy than in AUROC indicates that ranking
performance is preserved more strongly than hard-decision performance.

We note that $L_1$ logistic regression, which is also sparse and whose
coefficients are directly interpretable, matches or exceeds BIRDNet in
AUROC on five of the six datasets. 
We therefore do not claim a predictive
advantage over $L_1$ logistic regression. The two models instead provide
different explanatory objects: $L_1$ logistic regression assigns a scalar
weight to each feature independently, whereas each BIRDNet unit represents
a typed, directional relation between two features, and a prediction
decomposes into a hierarchy of such relations rather than a flat weighted
sum. We do not compare here which form of explanation is more useful in downstream
scientific analysis.

\begin{table}[!htb]
\caption{Classification performance (5-fold CV, mean$\pm$std).
  AUROC is one-vs-rest macro-averaged; leading zeros omitted.
  Best per dataset in bold. BIRDNet uses 3 to 95 times fewer active
  parameters than MatchedMLP (see Table~\ref{tab:sparsity}).}
\label{tab:results}
\centering
\small
\setlength{\tabcolsep}{2.5pt}
\begin{tabular}{lcccc}
\toprule
& BIRDNet & MatchedMLP & LogReg L1 & Random Forest \\
\midrule
\multicolumn{5}{l}{\emph{AUROC}} \\
UCI mice prot. & .998$\pm$.003 & \textbf{1.000$\pm$.000} & .999$\pm$.002 & 1.000$\pm$.000 \\
TCGA RPPA      & \textbf{.999$\pm$.000} & .999$\pm$.000 & .999$\pm$.000 & .999$\pm$.000 \\
GSE39582       & .958$\pm$.012 & \textbf{.976$\pm$.005} & .976$\pm$.005 & .972$\pm$.007 \\
UCI gene expr. & 1.000$\pm$.000 & 1.000$\pm$.000 & \textbf{1.000$\pm$.000} & 1.000$\pm$.000 \\
METABRIC       & .947$\pm$.009 & .960$\pm$.008 & .964$\pm$.005 & \textbf{.966$\pm$.005} \\
TCGA RNA-seq   & .996$\pm$.001 & .999$\pm$.000 & \textbf{1.000$\pm$.000} & .998$\pm$.000 \\
\midrule
\multicolumn{5}{l}{\emph{Accuracy}} \\
UCI mice prot. & .980$\pm$.018 & \textbf{.995$\pm$.006} & .983$\pm$.015 & .990$\pm$.009 \\
TCGA RPPA      & .966$\pm$.006 & .964$\pm$.004 & \textbf{.967$\pm$.002} & .956$\pm$.004 \\
GSE39582       & .756$\pm$.047 & \textbf{.822$\pm$.017} & .806$\pm$.034 & .811$\pm$.030 \\
UCI gene expr. & .988$\pm$.009 & \textbf{.996$\pm$.003} & .996$\pm$.005 & .995$\pm$.007 \\
METABRIC       & .753$\pm$.018 & .770$\pm$.026 & .792$\pm$.018 & \textbf{.798$\pm$.020} \\
TCGA RNA-seq   & .915$\pm$.011 & .959$\pm$.003 & \textbf{.970$\pm$.003} & .944$\pm$.007 \\
\bottomrule
\end{tabular}
\end{table}

\subsubsection*{Parameter efficiency.}

We report parameter counts in Table~\ref{tab:sparsity}, showing this for both BIRDNet and MLP. 
The four high-$d$
datasets saturate the total BIR-unit budget of $L \cdot h_{\max} = 10{,}000$
on every fold, while TCGA RPPA and UCI mice protein fall below the cap.
The classifier head, identical between BIRDNet and MatchedMLP, adds a
fixed overhead, so the compression ratio is bounded by the BIR-layer
contribution and approaches the $2/d$ asymptote of
Proposition~\ref{prop:sparsity}: $95\times$ on the four high-$d$ datasets
($d=2{,}000$), $32\times$ on TCGA RPPA ($d=258$), and $2.9\times$ on UCI
mice protein ($d=77$).

\begin{table}[!htb]
\caption{Parameter accounting per dataset (mean over folds).
  \emph{Width} is the total number of BIR units across all
  constructed layers; 
  \emph{BIR act.} counts active parameters in the BIR layers: two mask-allowed weights plus one bias per unit;
  \emph{Total act.} adds the dense classifier head;
  \emph{Ratio} = MatchedMLP / BIRDNet total active.}
\label{tab:sparsity}
\centering
\small
\setlength{\tabcolsep}{4pt}
\begin{tabular}{lrrrrr}
\toprule
Dataset           & Width  & BIR act. & Total act. & MatchedMLP & Ratio \\
\midrule
UCI mice protein  & 2.8k  &  8.5k & 186.8k &  544.1k &  2.9 \\
TCGA RPPA         & 7.1k  & 21.3k & 357.3k &  11.4M  & 31.8 \\
GSE39582          &  10k  & 30.0k & 370.5k &  35.4M  & 95.4 \\
UCI gene expr.    &  10k  & 30.0k & 370.4k &  35.4M  & 95.4 \\
METABRIC          &  10k  & 30.0k & 370.5k &  35.4M  & 95.4 \\
TCGA RNA-seq      &  10k  & 30.0k & 371.8k &  35.4M  & 95.1 \\
\bottomrule
\end{tabular}
\end{table}

\subsubsection*{Matched-topology control.}

To separate the effect of mined implications from degree-2 sparsity, we use
two controls. \emph{BIRDNet-Shuffle} permutes feature identities while
preserving the mined topology and BIR-type multiset.
\emph{BIRDNet-Random} samples the same number of feature pairs and types
uniformly, additionally destroying the degree distribution. Both use the
same training procedure as BIRDNet.
Table~\ref{tab:ablation} shows that Shuffle matches or exceeds BIRDNet on
five of six datasets (one-sided Wilcoxon signed-rank test, $W=1$,
$p=0.031$), while Random matches or improves upon Shuffle throughout.
The largest gains occur on GSE39582 and METABRIC.
In these experiments, the mined pairings do not improve predictive
performance over matched sparse controls. Their contribution is instead
symbolic: shuffled or random pairs do not correspond to mined implications
and therefore do not support the propositional interpretation and rule
extraction used in Section~\ref{sec:rules}.

\begin{table}[!htb]
\caption{Matched-topology controls (5-fold CV, mean AUROC).
\emph{Shuffle} permutes feature identities while preserving the mined
topology and BIR-type multiset; \emph{Random} additionally destroys the
degree distribution.}
\label{tab:ablation}
\centering
\small
\setlength{\tabcolsep}{5pt}
\begin{tabular}{lccc}
\toprule
Dataset & BIRDNet & Shuffle & Random \\
\midrule
UCI mice protein & 0.998 & 0.999 & 1.000 \\
TCGA RPPA        & 0.999 & 0.999 & 0.999 \\
GSE39582         & 0.958 & 0.978 & 0.982 \\
UCI gene expr.   & 1.000 & 1.000 & 1.000 \\
METABRIC         & 0.947 & 0.962 & 0.970 \\
TCGA RNA-seq     & 0.996 & 0.998 & 0.999 \\
\bottomrule
\end{tabular}
\end{table}

\subsubsection*{Label-free preselection control.}
We repeat METABRIC using marginal variance instead of the label-based F-test.
The main rules remain largely unchanged, and AUROC increases slightly from
$0.947\pm0.009$ to $0.959\pm0.007$. The BIRDNet--Shuffle gap also narrows
from $0.015$ to $0.002$ AUROC. On METABRIC, this suggests that the recovered
structure is not solely driven by supervised preselection.

\subsubsection{Explanatory evaluation}
\label{sec:rules}

\subsubsection*{Per-class explanation.}
We extract per-class rules from BIRDNet's first BIR layer using a final
80/20 split (seed 42) of each dataset. This provides a clean held-out set
for estimating rule precision and lift, separate from the 5-fold CV used
for predictive evaluation. Across all six datasets, the highest-precision
rules are dominated by co-expression ($T_0$) and co-repression ($T_1$),
although all six BIR types are considered during implication mining.
Table~\ref{tab:rules} reports the top rule for representative classes from
METABRIC and TCGA RNA-seq.

\begin{table}[!htb]
\caption{Top extracted rule per class on held-out test data.
  $A$ denotes \texttt{high}($\cdot$), and $\neg A$ denotes
  \texttt{low}($\cdot$). \emph{Prec.} = $\Pr(c \mid \text{unit active})$;
  \emph{Lift} = $\Pr(c \mid \text{unit active})/\Pr(c)$.}
\label{tab:rules}
\centering
\small
\setlength{\tabcolsep}{4pt}
\begin{tabular}{llrr}
\toprule
Class & Rule & Prec. & Lift \\
\midrule
\multicolumn{4}{l}{\emph{METABRIC (breast cancer subtypes)}} \\
HER2          & $\textit{STARD3} \to \textit{GRB7}$             & 0.55 & 4.9 \\
Basal         & $\textit{ROPN1B} \to \textit{ROPN1}$            & 0.64 & 6.1 \\
LumA          & $\neg\textit{UBE2C} \to \neg\textit{AURKA}$     & 0.73 & 2.1 \\
Claudin-low   & $\textit{IL23A} \to \textit{CCL5}$              & 0.56 & 5.0 \\
\midrule
\multicolumn{4}{l}{\emph{TCGA RNA-seq (pan-cancer)}} \\
Brain LGG       & $\textit{HEPN1} \to \textit{HEPACAM}$         & 0.49 & 9.4 \\
Colon Adeno.    & $\textit{CEACAM5} \to \textit{FXYD3}$         & 0.25 & 7.7 \\
Breast Invasive & $\textit{SPDEF} \to \textit{TMC5}$            & 0.46 & 3.8 \\
Cervical        & $\textit{KRT6A} \to \textit{KRT15}$           & 0.17 & 5.6 \\
\bottomrule
\end{tabular}
\end{table}

The METABRIC rules are consistent with known subtype biology.
For HER2, \textit{STARD3} and \textit{GRB7} lie within the 17q12
\textit{ERBB2} amplicon and are frequently co-amplified in HER2-positive
breast cancer~\cite{kauraniemi2006activation}; notably, all five top-ranked
HER2 rules pair genes from this region. The Basal rule links
\textit{ROPN1B} and \textit{ROPN1}, associated with the basal
phenotype~\cite{nielsen2004immunohistochemical}. The LumA rule
$\neg\textit{UBE2C} \to \neg\textit{AURKA}$ captures low expression of
proliferation-related genes, consistent with the lower-proliferation
character of LumA tumours~\cite{parker2009supervised}. Finally,
\textit{IL23A} $\to$ \textit{CCL5} is consistent with the prominent
immune-infiltration signature of claudin-low tumours~\cite{prat2010phenotypic}.

Similarly, on TCGA RNA-seq, several top rules are consistent with
tissue-specific expression patterns. Lung Adenocarcinoma is identified by
\textit{SFTPA1}/\textit{SFTPA2}, and all five of its top-ranked rules pair
genes of the alveolar surfactant system~\cite{whitsett2010alveolar}. Other
high-lift pairs, such as \textit{CEACAM5}/\textit{FXYD3} for Colon
Adenocarcinoma, \textit{KRT6A}/\textit{KRT15} for Cervical cancer, and
\textit{HEPN1}/\textit{HEPACAM} for Brain LGG, are reported as mined
regularities; we do not attempt independent biological validation here.

\subsubsection*{Per-instance explanation}

For a held-out TCGA RNA-seq sample correctly classified as Head \& Neck
Squamous Cell Carcinoma (HNSC), layer-wise relevance propagation identifies
the following path:
\begin{equation*}
\begin{split}
\underbrace{\textit{KRT6A} \to \textit{KRT16}}_{\text{layer-0 unit}}
\;\rightsquigarrow\;
&\underbrace{\text{BIR}(u_{15},u_{1801})}_{\text{layer-1 unit}}
\;\rightsquigarrow\;
\underbrace{z_{13}}_{\text{head}}\\
&\rightsquigarrow\;
\text{class}=\text{HNSC}~(1.00).
\end{split}
\end{equation*}
Both keratins are strongly expressed in this sample (standardised values
$2.04$ and $2.28$), consistent with the squamous phenotype. The same
layer-0 rule $u_{15}$ is also reused by another layer-1 unit together with
a rule involving \textit{KRT16} and \textit{SFN}. A parallel branch
contains another shared rule contributing \textit{CEACAM6},
\textit{SFN}, and \textit{LAMC2}. Thus, the evidence for the prediction
is not confined to a single feature pair: several related
squamous-epithelial rules contribute through shared intermediate units.
Because layer-0 rules can feed more than one higher-level unit, the
explanation is naturally a directed acyclic graph rather than a single
chain. Layer-0 units refer directly to pairs of input features, whereas a
layer-1 unit represents a relation between two layer-0 rules. Recovering a
gene-level interpretation therefore requires tracing the hierarchy back to
the inputs. This loss of immediate legibility remains limited in our
experiments, since all six datasets yield networks of depth at most $L=2$.

\section{Concluding Remarks}
\label{sec:concl}

BIRDNet uses mined Boolean implications to build highly sparse networks
with a traceable symbolic structure. Across six scientific datasets,
BIRDNet remains within $0.02$ AUROC of the strongest baseline while using
$3\times$--$95\times$ fewer active parameters than the matched dense MLP.
The corresponding accuracy gaps are larger on some datasets, which we
report explicitly. Unlike post-hoc explanations or surrogate models,
BIRDNet predictions can be traced through a hierarchy of named
propositional relations back to the input features.

\subsubsection*{Applicability and limitations.}
BIRDNet is most applicable when features admit meaningful binarisation,
the feature dimension is sufficiently large for degree-2 sparsity to yield
substantial compression, and the sample size provides adequate power for
the implication test at $p^*=10^{-6}$. Transcriptomic and proteomic data
satisfy these conditions in our experiments.

The method has three main limitations. First, the current implementation
uses only pairwise implications and may miss higher-order relationships.
Second, the structure is learned entirely from data and does not incorporate
prior domain knowledge, which may be important in data-limited scientific
settings~\cite{dash2022review,dash2022thesis}. Third, our ablation experiments
show that the mined topology incurs a predictive cost relative to matched
sparse controls. This highlights a trade-off between symbolic
interpretability and predictive performance~\cite{rudin2019stop}. In future
work, we will study how implication mining can scale to substantially larger
feature spaces and datasets, and examine richer structural priors that retain
the symbolic interpretation while reducing the predictive gap observed here.

\begin{acks}
The author thanks Debashis Sahoo (UC San Diego) for helpful discussions on Boolean implication networks and for comments on an earlier version of this work. 
\end{acks}

\section*{GenAI Usage Disclosure}

We used Claude Opus as a generative AI assistant for English editing, interpreting selected BIRDNet rules, and debugging the open-source implementation released with the paper. All scientific contributions, including the method formulation, experimental design, analyses, and conclusions, are solely the work of the author.

\bibliographystyle{ACM-Reference-Format}
\bibliography{refs}

@article{sahoo2008,
  title={Boolean implication networks derived from large scale, whole genome microarray datasets},
  author={Sahoo, Debashis and Dill, David L and Gentles, Andrew J and Tibshirani, Robert and Plevritis, Sylvia K},
  journal={Genome biology},
  volume={9},
  number={10},
  pages={R157},
  year={2008},
  publisher={Springer}
}

@article{sahoo2012power,
  title={The power of boolean implication networks},
  author={Sahoo, Debashis},
  journal={Frontiers in Physiology},
  volume={3},
  pages={276},
  year={2012},
  publisher={Frontiers Research Foundation}
}

@article{ma2018dcell,
  title={Using deep learning to model the hierarchical structure and function of a cell},
  author={Ma, Jianzhu and Yu, Michael Ku and Fong, Samson and Ono, Keiichiro and Sage, Eric and Demchak, Barry and Sharan, Roded and Ideker, Trey},
  journal={Nature methods},
  volume={15},
  number={4},
  pages={290--298},
  year={2018},
  publisher={Nature Publishing Group US New York}
}

@article{elmarakeby2021pnet,
  title={Biologically informed deep neural network for prostate cancer discovery},
  author={Elmarakeby, Haitham A and Hwang, Justin and Arafeh, Rand and Crowdis, Jett and Gang, Sydney and Liu, David and AlDubayan, Saud H and Salari, Keyan and Kregel, Steven and Richter, Camden and others},
  journal={Nature},
  volume={598},
  number={7880},
  pages={348--352},
  year={2021},
  publisher={Nature Publishing Group UK London}
}

@article{srinivasan2024crm,
  title={Composition of relational features with an application to explaining black-box predictors},
  author={Srinivasan, Ashwin and Baskar, A and Dash, Tirtharaj and Shah, Devanshu},
  journal={Machine Learning},
  volume={113},
  number={3},
  pages={1091--1132},
  year={2024},
  publisher={Springer}
}

@article{bach2015lrp,
  title={On pixel-wise explanations for non-linear classifier decisions by layer-wise relevance propagation},
  author={Bach, Sebastian and Binder, Alexander and Montavon, Gr{\'e}goire and Klauschen, Frederick and M{\"u}ller, Klaus-Robert and Samek, Wojciech},
  journal={PloS one},
  volume={10},
  number={7},
  pages={e0130140},
  year={2015},
  publisher={Public Library of Science San Francisco, CA USA}
}

@article{curtis2012genomic,
  title={The genomic and transcriptomic architecture of 2,000 breast tumours reveals novel subgroups},
  author={Curtis, Christina and Shah, Sohrab P and Chin, Suet-Feung and Turashvili, Gulisa and Rueda, Oscar M and Dunning, Mark J and Speed, Doug and Lynch, Andy G and Samarajiwa, Shamith and Yuan, Yinyin and others},
  journal={Nature},
  volume={486},
  number={7403},
  pages={346--352},
  year={2012},
  publisher={Nature Publishing Group UK London}
}

@article{weinstein2013cancer,
  title={The cancer genome atlas pan-cancer analysis project},
  author={Weinstein, John N and Collisson, Eric A and Mills, Gordon B and Shaw, Kenna R and Ozenberger, Brad A and Ellrott, Kyle and Shmulevich, Ilya and Sander, Chris and Stuart, Joshua M},
  journal={Nature genetics},
  volume={45},
  number={10},
  pages={1113--1120},
  year={2013},
  publisher={Nature Publishing Group}
}

@article{higuera2015self,
  title={Self-organizing feature maps identify proteins critical to learning in a mouse model of down syndrome},
  author={Higuera, Clara and Gardiner, Katheleen J and Cios, Krzysztof J},
  journal={PloS one},
  volume={10},
  number={6},
  pages={e0129126},
  year={2015},
  publisher={Public Library of Science San Francisco, CA USA}
}

@inproceedings{xia2022structured,
    title = "Structured Pruning Learns Compact and Accurate Models",
    author = "Xia, Mengzhou  and
      Zhong, Zexuan  and
      Chen, Danqi",
    editor = "Muresan, Smaranda  and
      Nakov, Preslav  and
      Villavicencio, Aline",
    booktitle = "Proceedings of the 60th Annual Meeting of the Association for Computational Linguistics (Volume 1: Long Papers)",
    month = may,
    year = "2022",
    address = "Dublin, Ireland",
    publisher = "Association for Computational Linguistics",
    url = "https://aclanthology.org/2022.acl-long.107/",
    doi = "10.18653/v1/2022.acl-long.107",
    pages = "1513--1528"
}

@article{dash2022review,
  title={A review of some techniques for inclusion of domain-knowledge into deep neural networks},
  author={Dash, Tirtharaj and Chitlangia, Sharad and Ahuja, Aditya and Srinivasan, Ashwin},
  journal={Scientific Reports},
  volume={12},
  number={1},
  pages={1040},
  year={2022},
  publisher={Nature Publishing Group UK London}
}

@article{kauraniemi2006activation,
  title={Activation of multiple cancer-associated genes at the ERBB2 amplicon in breast cancer},
  author={Kauraniemi, P and Kallioniemi, A},
  journal={Endocrine-related cancer},
  volume={13},
  number={1},
  pages={39--49},
  year={2006},
  publisher={BioScientifica}
}

@article{nielsen2004immunohistochemical,
  title={Immunohistochemical and clinical characterization of the basal-like subtype of invasive breast carcinoma},
  author={Nielsen, Torsten O and Hsu, Forrest D and Jensen, Kristin and others},
  journal={Clinical cancer research},
  volume={10},
  number={16},
  pages={5367--5374},
  year={2004},
  publisher={American Association for Cancer Research}
}

@article{parker2009supervised,
  title={Supervised risk predictor of breast cancer based on intrinsic subtypes},
  author={Parker, Joel S and Mullins, Michael and Cheang, Maggie CU and others},
  journal={Journal of clinical oncology},
  volume={27},
  number={8},
  pages={1160--1167},
  year={2009},
  publisher={American Society of Clinical Oncology}
}

@article{garcez2023neurosymbolic,
  title={Neurosymbolic ai: The 3 rd wave},
  author={Garcez, Artur d’Avila and Lamb, Luis C},
  journal={Artificial Intelligence Review},
  volume={56},
  number={11},
  pages={12387--12406},
  year={2023},
  publisher={Springer}
}

@article{wang2024towards,
  title={Towards data-and knowledge-driven AI: a survey on neuro-symbolic computing},
  author={Wang, Wenguan and Yang, Yi and Wu, Fei},
  journal={IEEE transactions on pattern analysis and machine intelligence},
  volume={47},
  number={2},
  pages={878--899},
  year={2024},
  publisher={IEEE}
}

@article{srinivasan2019logical,
  title={Logical explanations for deep relational machines using relevance information},
  author={Srinivasan, Ashwin and Vig, Lovekesh and Bain, Michael},
  journal={Journal of Machine Learning Research},
  volume={20},
  number={130},
  pages={1--47},
  year={2019}
}

@article{whitsett2010alveolar,
  title={Alveolar surfactant homeostasis and the pathogenesis of pulmonary disease},
  author={Whitsett, Jeffrey A and Wert, Susan E and Weaver, Timothy E},
  journal={Annual review of medicine},
  volume={61},
  number={1},
  pages={105--119},
  year={2010},
  publisher={Annual Reviews}
}

@article{prat2010phenotypic,
  title={Phenotypic and molecular characterization of the claudin-low intrinsic subtype of breast cancer},
  author={Prat, Aleix and Parker, Joel S and Karginova, Olga and Fan, Cheng and Livasy, Chad and Herschkowitz, Jason I and He, Xiaping and Perou, Charles M},
  journal={Breast cancer research},
  volume={12},
  number={5},
  pages={R68},
  year={2010},
  publisher={Springer}
}

@misc{fiorini2016uci,
  author       = {Fiorini, Samuele},
  title        = {{gene expression cancer RNA-Seq}},
  year         = {2016},
  howpublished = {UCI Machine Learning Repository},
  note         = {{DOI}: https://doi.org/10.24432/C5R88H}
}

@article{akbani2014pan,
  title={A pan-cancer proteomic perspective on The Cancer Genome Atlas},
  author={Akbani, Rehan and Ng, Patrick Kwok Shing and Werner, Henrica MJ and Shahmoradgoli, Maria and Zhang, Fan and Ju, Zhenlin and Liu, Wenbin and Yang, Ji-Yeon and Yoshihara, Kosuke and Li, Jun and others},
  journal={Nature communications},
  volume={5},
  number={1},
  pages={3887},
  year={2014},
  publisher={Nature Publishing Group UK London}
}

@article{marisa2013gene,
  title={Gene expression classification of colon cancer into molecular subtypes: characterization, validation, and prognostic value},
  author={Marisa, Laetitia and de Reyni{\`e}s, Aur{\'e}lien and Duval, Alex and Selves, Janick and Gaub, Marie Pierre and Vescovo, Laure and Etienne-Grimaldi, Marie-Christine and Schiappa, Renaud and Guenot, Dominique and Ayadi, Mira and others},
  journal={PLoS medicine},
  volume={10},
  number={5},
  pages={e1001453},
  year={2013},
  publisher={Public Library of Science San Francisco, USA}
}

@inproceedings{frankle2019lottery,
  title={The Lottery Ticket Hypothesis: Finding Sparse, Trainable Neural Networks},
  author={Frankle, Jonathan and Carbin, Michael},
  booktitle={International Conference on Learning Representations (ICLR)},
  year={2019},
  publisher={OpenReview.net},
  address={New Orleans, LA, USA},
  pages={1--42},
  url={https://openreview.net/forum?id=rJl-b3RcF7}
}

@inproceedings{agrawal1994fast,
author = {Agrawal, Rakesh and Srikant, Ramakrishnan},
title = {Fast Algorithms for Mining Association Rules in Large Databases},
year = {1994},
isbn = {1558601538},
publisher = {Morgan Kaufmann Publishers Inc.},
address = {San Francisco, CA, USA},
booktitle = {Proceedings of the 20th International Conference on Very Large Data Bases},
pages = {487–499},
numpages = {13},
series = {VLDB '94}
}

@article{han2000mining,
  title={Mining frequent patterns without candidate generation},
  author={Han, Jiawei and Pei, Jian and Yin, Yiwen},
  journal={ACM SIGMOD record},
  volume={29},
  number={2},
  pages={1--12},
  year={2000},
  publisher={ACM New York, NY, USA}
}

@phdthesis{dash2022thesis,
  title={Inclusion of Symbolic Domain-Knowledge into Deep Neural Networks},
  author={Dash, Tirtharaj},
  year={2022},
  school={BIRLA INSTITUTE OF TECHNOLOGY AND SCIENCE PILANI}
}

@article{rudin2019stop,
  title={Stop explaining black box machine learning models for high stakes decisions and use interpretable models instead},
  author={Rudin, Cynthia},
  journal={Nature machine intelligence},
  volume={1},
  number={5},
  pages={206--215},
  year={2019},
  publisher={Nature Publishing Group UK London}
}

\end{document}